\documentclass[sigconf]{acmart}

\AtBeginDocument{%
  \providecommand\BibTeX{{%
    \normalfont B\kern-0.5em{\scshape i\kern-0.25em b}\kern-0.8em\TeX}}}

\copyrightyear{2020}
\acmYear{2020}
\setcopyright{acmcopyright}
\acmConference[SIGIR '20] {43rd International ACM SIGIR Conference on Research and Development in Information Retrieval}{July 25--30, 2020}{Virtual Event, China}
\acmBooktitle{43rd International ACM SIGIR Conference on Research and Development in Information Retrieval (SIGIR '20), July 25--30, 2020, Virtual Event, China}
\acmPrice{15.00}
\acmDOI{10.1145/3397271.3401039}
\acmISBN{978-1-4503-8016-4/20/07}

\usepackage{amsmath}
\usepackage{amssymb}
\usepackage{multirow}
\usepackage{array}
\usepackage{algorithm}
\usepackage{algorithmic}
\usepackage{enumitem}
\usepackage{comment}

\begin{document}
\fancyhead{}
\title{A Unified Dual-view Model for Review Summarization and Sentiment Classification with Inconsistency Loss}

\author{Hou Pong Chan}
\authornote{Both authors contributed equally to this research.}
\email{hpchan@cse.cuhk.edu.hk}
\affiliation{
  \institution{The Chinese University of Hong Kong}
  \city{Shatin, N.T., Hong Kong}
}

\author{Wang Chen}
\authornotemark[1]
\email{wchen@cse.cuhk.edu.hk}
\affiliation{
  \institution{The Chinese University of Hong Kong}
  \city{Shatin, N.T., Hong Kong}
}

\author{Irwin King}
\email{king@cse.cuhk.edu.hk}
\affiliation{%
  \institution{The Chinese University of Hong Kong}
  \city{Shatin, N.T., Hong Kong}
}



\renewcommand{\shortauthors}{Trovato and Tobin, et al.}
\newcommand{\red}[1]{\textcolor{red}{{#1}}}
\newcommand{\blue}[1]{\textcolor{blue}{{#1}}}

\begin{abstract}
Acquiring accurate summarization and sentiment from user reviews is an essential component of modern e-commerce platforms. Review summarization aims at generating a concise summary that describes the key opinions and sentiment of a review, while sentiment classification aims to predict a sentiment label indicating the sentiment attitude of a review. To effectively leverage the shared sentiment information in both review summarization and sentiment classification tasks, we propose a novel dual-view model that jointly improves the performance of these two tasks. In our model, an encoder first learns a context representation for the review, then a summary decoder generates a review summary word by word. After that, a source-view sentiment classifier uses the encoded context representation to predict a sentiment label for the review, while a summary-view sentiment classifier uses the decoder hidden states to predict a sentiment label for the generated summary. During training, we introduce an inconsistency loss to penalize the disagreement between these two classifiers. It helps the decoder to generate a summary to have a consistent sentiment tendency with the review and also helps the two sentiment classifiers learn from each other. Experiment results on four real-world datasets from different domains demonstrate the effectiveness of our model. 
\end{abstract}

\ccsdesc[500]{Information systems~Summarization}
\ccsdesc[500]{Information systems~Sentiment analysis}
\ccsdesc[300]{Information systems~Recommender systems}

\keywords{Review Summarization; Sentiment Classification; Multi-task Learning; Text Summarization}

\maketitle

\section{Introduction}
User reviews are precious for companies to improve the quality of their services or products since reviews express the opinions of their customers. Nowadays, a company can easily collect reviews from users via e-commerce platforms and recommender systems, but it is difficult to read through all the wordy user reviews. Therefore, distilling salient information from user reviews is necessary. To achieve such a goal, review summarization and sentiment classification are continuously explored by plenty of researchers. 
The objective of review summarization is to generate a short and concise summary that expresses the key opinions and sentiment of the review. 
Sentiment classification is the task of predicting the sentiment label which indicates the sentiment attitude of the review. For example, a sentiment label ranges from 1 to 5, where 1 indicates the most negative attitude and 5 indicates the most positive attitude. 

\begin{figure}[t]
\begin{tabular}{|p{0.95\columnwidth}|}
\hline
\textbf{Review:} very colorful and decent for prep and serving small kids. bright colors are nice and food colors don't seem to transfer to the bowls. we use them most frequently for salad dressing when the kids have baby carrots or celery sticks. my complaint is that the bases are too small and consequently they tip over very easily. that's ok for ranch dressing or ketchup, not so good for soy sauce. if you flip them inside out it gives them a wider base which makes them more stable. since they flex they make their own funnel 
... we have some stainless mini bowls which are almost an inch larger in diameter (but shallower) which makes them a little more handy ... \\
\hline
\textbf{Summary:} very colorful and handy, but tippy and smaller than expected. \\ \hline
\textbf{Sentiment label:} 3 out of 5                                                \\ \hline
\end{tabular}
\caption{An example of truncated review and its corresponding summary and sentiment label.
}\label{fig:intro-example}
\vspace{-0.1in}
\end{figure}

Figure~\ref{fig:intro-example} shows an example of a review with its summary and sentiment label. 
As shown in the summary, the user thinks the product is ``colorful and handy, but tippy and smaller than expected''. The sentiment label of the review is 3 out of 5, indicating the review has a neutral sentiment attitude. 
We can observe that the sentiment tendency of the review summary is consistent with the sentiment label of the review, which indicates there exists a close tie between these two tasks~\cite{DBLP:conf/ijcai/MaSLR18}.
Hence, they can benefit each other. Specifically, the sentiment classification task can guide a summarization model to capture the sentiment attitude of the review. Meanwhile, the review summarization task can remove the less-informative information from the review, which can assist a sentiment classification model to predict the sentiment label of a review. 

However, most existing methods can only solve either review summarization or sentiment classification problem. For review summarization, traditional extractive methods select important phrases or sentences from the input review text to produce an output summary. Recent abstractive methods for review summarization adopt the attentional encoder-decoder framework~\cite{DBLP:conf/iclr/BahdanauCB14} to generate an output summary word by word, which can generate novel words that do not appear in the review and produce a more concise summary. For sentiment classification, early methods are based on support vector machines or statistical models, while recent methods employ recurrent neural network (RNN)~\cite{DBLP:journals/neco/HochreiterS97}. 
Though some previous methods~\cite{mane2015summarization,hole2013real} can predict both the sentiment label and the summary for a social media text, the sentiment classification and summarization modules are trained separately and they rely on rich hand-crafted features. 

Recently, Ma et al.~\cite{DBLP:conf/ijcai/MaSLR18} proposed an end-to-end model that jointly improves the performance of review summarization and sentiment classification. 
In their model, an encoder first learns a context representation for the review, which captures the context and sentiment information. Based on this representation, the decoder iteratively computes a hidden state and uses it to generate a summary word from a predefined vocabulary until the end-of-sequence (EOS) token is generated. 
Meanwhile, an attention layer utilizes the decoder hidden states to attend the review to compute a weighted sum of the review context representation, which acts as a summary-aware context representation of the review. 
Next, the review context representation and the summary-aware context representation are fed to a max-pooling based classifier to predict the sentiment label. However, both of these representations collect sentiment information from the review only. Thus, the model does not fully utilize the sentiment information existing in the summary. 

To effectively leverage the sentiment information in the review and the summary, we propose a dual-view model for joint review summarization and sentiment classification. 
In our model, the sentiment information in the review and the summary are modeled by the source-view and summary-view sentiment classifiers respectively. 
The source-view sentiment classifier uses the review context representation from the encoder to predict a sentiment label for the review, while the summary-view sentiment classifier utilizes the decoder hidden states to predict a sentiment label for the generated summary. 
The ground-truth sentiment label of the review will be used to compute a classification loss for both of these classifiers. 
In addition, we also introduce an inconsistency loss function to penalize the disagreement between these two classifiers. 

By promoting the consistency between the source-view and summary-view classifiers, we encourage the sentiment information in the decoder states to be close to that in the review context representation, which helps the decoder to generate a summary that has the same sentiment attitude as the review. 
Moreover, the source-view and summary-view sentiment classifiers can learn from each other to improve the sentiment classification performance. This shares a similar spirit with the multi-view learning paradigm in semi-supervised learning~\cite{DBLP:journals/corr/multi-view-survey}. 
Our model, therefore, provides more supervision signals for both review summarization and sentiment classification without additional training data. 
During testing, we use the sentiment label predicted by the source-view classifier as the final classification output since the summary-view sentiment classifier will be affected by the exposure bias issue~\cite{DBLP:journals/corr/RanzatoCAZ15} in the testing stage (more details in Section~\ref{sec:dual-view-classifier}). 

We conduct extensive experiments to evaluate the performance of our dual-view model. Experiment results on four real-world datasets from different domains demonstrate that our model outperforms strong baselines on both review summarization and sentiment classification. 
The ablation study shows the effectiveness of each individual component of our model. 
Furthermore, we also compare the classification results of our source-view sentiment classifier, summary-view sentiment classifier, and a merged sentiment classifier that combines the former two classifiers.

We summarize our main contributions as follows: (1) we propose a novel dual-view model for jointly improving the performance of review summarization and sentiment classification; (2) we introduce an inconsistency loss to penalize the inconsistency between our source-view and summary-view sentiment classifiers, which benefits both review summarization and sentiment classification; and (3) experimental results on benchmark datasets show that our model outperforms the state-of-the-art models for the joint review summarization and sentiment classification task. 

\section{Related Work}
\noindent \textbf{Opinion Summarization. }
Review summarization belongs to the area of opinion summarization~\cite{DBLP:conf/coling/GanesanZH10,DBLP:conf/emnlp/GeraniMCNN14,DBLP:conf/aaai/LiLZ19,frermann-klementiev-2019-inducing_fromReviewer,DBLP:conf/cikm/TianY019_fromReviewer, DBLP:conf/recsys/MustoRGLS19_fromReviewer}. 
Early approaches for opinion summarization are extractive, i.e., they can only produce words that appear in the input document. Ganesan et al.~\cite{DBLP:conf/coling/GanesanZH10} proposed a graph-based approach. It first converts the input opinion text into a directed graph. Then it applies heuristic rules to score different paths that encode valid sentences and takes the top-ranked paths as the output summary. Hu and Liu~\cite{DBLP:conf/kdd/HuL04} proposed a method that first identifies product features mentioned in the reviews and then extracts opinion sentences for the identified features. An unsupervised learning method is proposed to extract a review summary by exploiting the helpfulness scores in reviews~\cite{DBLP:conf/coling/XiongL14}. 

On the other hand, abstractive approaches for opinion summarization can generate novel words that do not appear in the input document. Gerani et al.~\cite{DBLP:conf/emnlp/GeraniMCNN14} proposed a template filling strategy to generate a review summary. Wang and Ling~\cite{DBLP:conf/naacl/WangL16} applied the attentional encoder-decoder model to generate an abstractive summary for opinionated documents. 
All of the above methods consider opinion summarization in multiple documents setting, while our work considers opinion summarization on single document setting. Li et al.~\cite{DBLP:conf/aaai/LiLZ19} studied the problem of personalized review summarization on single review setting. They incorporated a user's frequently used words into the encoder-decoder model to generate a user-aware summary. In contrast, we focus on modeling the shared sentiment information between the tasks of review summarization and sentiment classification, which is orthogonal to the personalized review generation problem. Hsu et al.~\cite{DBLP:conf/acl/SunHLLMT18} proposed a unified model for extractive and abstractive summarization with an inconsistency loss to penalize the disagreement between the extractive and abstractive attention scores. Compared to their model, which penalizes the inconsistencies between two different attention methods on the same view (i.e., the source input), we introduce an inconsistency loss to penalize the outputs of two different classifiers on different views (i.e., the source view and the summary view). 

\noindent \textbf{Review Sentiment Classification. }
Review sentiment classification aims at analyzing online consumer reviews and predicting the sentiment attitude of a consumer towards a product. 
The review sentiment classification tasks can be categorized into three groups: document-level, sentence-level, and aspect-level sentiment classification~\cite{ZhangLei@18_sentiment_survey,DBLP:conf/cikm/ChengZZKZW17_jiajun}. We focus on document-level sentiment classification that is to assign an overall sentiment orientation to the input document~\cite{ZhangLei@18_sentiment_survey}, which is usually treated as a kind of document classification task~\cite{DBLP:conf/emnlp/PangLV02,DBLP:conf/acl/PangL05}. Traditional methods focus on designing effective features that are used in either supervised learning methods~\cite{DBLP:conf/coling/QuIW10,DBLP:conf/ijcnlp/Gao0KK13} or graph-based semi-supervised methods~\cite{goldberg-zhu-2006-seeing}. Recently, neural network based methods which do not require hand-crafted features achieve state-of-the-art performance on this task. For example, Tang et al.~\cite{tang-etal-2015-document} proposed a neural network based hierarchical encoding process to learn an effective review representation. 
Hierarchical attention mechanisms~\cite{yang-etal-2016-hierarchical,yin-etal-2017-document,zhou-etal-2016-attention-based} are also extensively explored in this task for constructing an effective representation of the review. Different from previous methods that are designed only for the review sentiment classification task, we propose a unified model for simultaneously generating the summary of the review and classifying the review sentiment. 

\noindent \textbf{Joint Text Summarization and Classification. }
There are different joint models for both text summarization and classification. 
Cao et al.~\cite{DBLP:conf/aaai/CaoLLW17} proposed a neural network model that jointly classifies the category and extracts summary sentences for a group of news articles, but it can only improve the performance of text summarization. 
Yang et al.~\cite{DBLP:conf/coling/YangQSLZZ18} proposed a joint model that uses domain classification as an auxiliary task to improve the performance of review summarization. Moreover, their model uses different lexicons to find out sentiment words and aspect words from the review text, and then incorporates them into the decoder via attention mechanism to generate aspect/sentiment-aware review summaries. The above two methods use a domain classification task to improve the performance of summarization, while our method jointly improves the performance of both review summarization and sentiment classification. 
Two models~\cite{mane2015summarization,hole2013real} were proposed to jointly extract a summary and predict the sentiment label for a social media post, but the summarization module and classification module of these models are trained separately and they require rich hand-crafted features. 
Recently, Ma et al.~\cite{DBLP:conf/ijcai/MaSLR18} proposed an end-to-end neural model that jointly improves the performance of the review summarization and sentiment classification tasks. 
However, their sentiment classifier only collects sentiment information from the review. 
Our dual-view model has a source-view sentiment classifier and a summary-view sentiment classifier to model the sentiment information in the review and the summary respectively. We also introduce an inconsistency loss to encourage the consistency between these two classifiers.

\section{Preliminary}

\noindent \textbf{Notations.} 
We use bold lowercase characters to denote vectors, bold upper case characters to denote matrices and calligraphy characters to denote sets. We use $\mathbf{W}$ and $\mathbf{b}$ to denote a projection matrix and a bias vector in a neural network layer. 

\noindent \textbf{Problem definition.} We formally define the problem of review summarization and sentiment classification as follows. 
Given a review text $\mathbf{x}$, we output the summary $\mathbf{y}$ and sentiment label $z$ of the review text. The review text $\mathbf{x}$ and its summary $\mathbf{y}$ are sequences of words, i.e., $\mathbf{x}=[x_{1},\ldots,x_{L_{\mathbf{x}}}]$ and $\mathbf{y}=[y_{1},\ldots,y_{L_{\mathbf{y}}}]$, where $L_{\mathbf{x}}$ and $L_{\mathbf{y}}$ denotes the numbers of word in $\mathbf{x}$ and $\mathbf{y}$ respectively. The sentiment label $z\in \{1,2,\ldots, K\}$ is an integer that indicates the sentiment attitude of the review text, where $1$ denotes the most negative sentiment and $K$ denotes the most positive sentiment. 

\section{Model Architecture}

\subsection{Overview}
\begin{figure}[t]
  \centering
  \includegraphics[width=\linewidth]{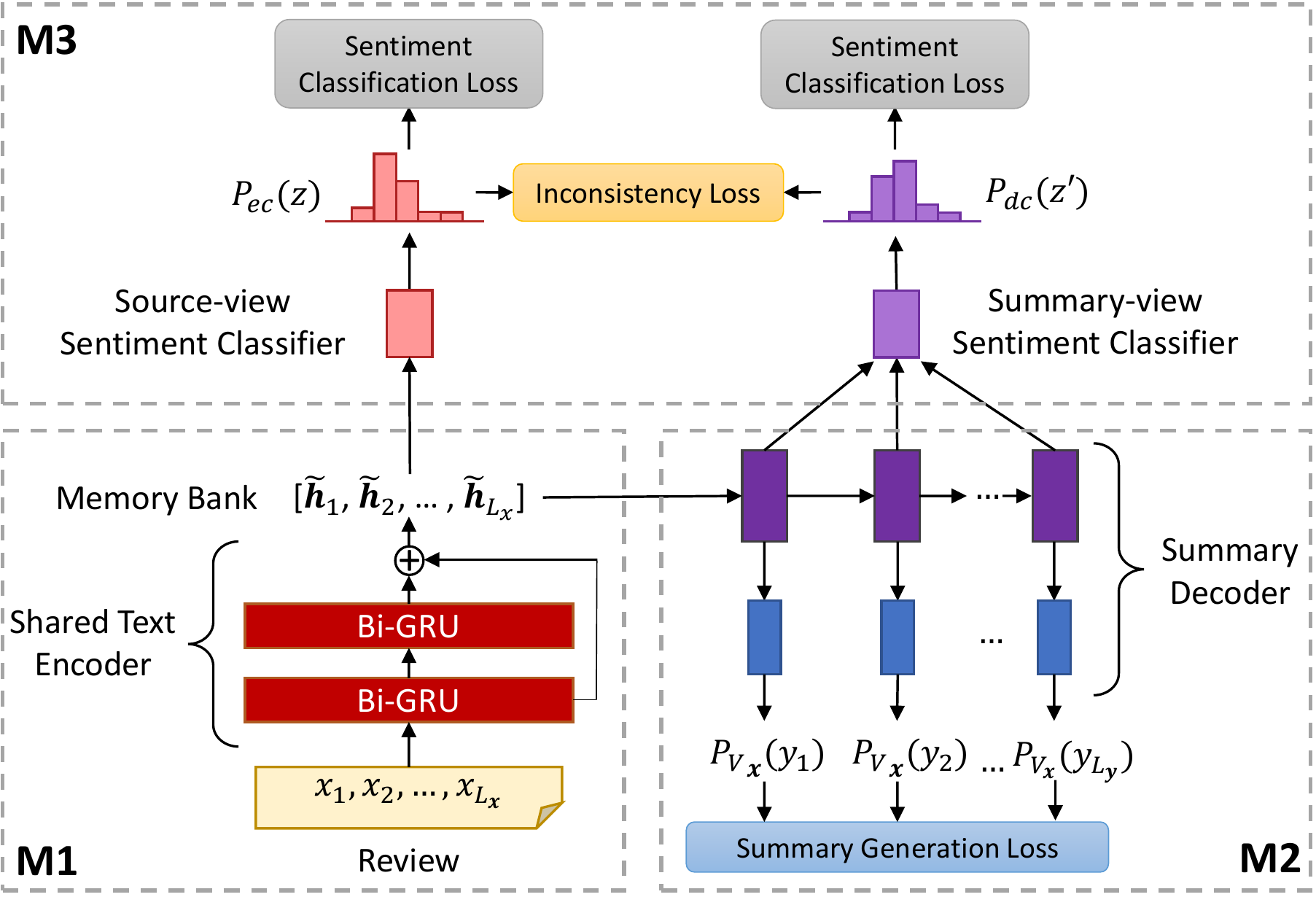}
  \caption{Overall model architecture consisting of (M1) shared text encoder, (M2) summary decoder, and (M3) dual-view sentiment classification module. The shared text encoder converts the input review text into a memory bank. Based on the memory bank, the summary decoder generates the review summary word by word and receives a summary generation loss. 
  The source-view (summary-view) sentiment classifier uses the memory bank (hidden states) from the encoder (decoder) to predict a sentiment label for the review (summary) and it receives a sentiment classification loss. 
  An inconsistency loss is applied to penalize the disagreement between the source-view and summary-view sentiment classifiers. }
  \label{fig:architecture}
\end{figure}

Our joint model consists of three major modules: (M1) shared text encoder module, (M2) summary decoder module, and (M3) dual-view sentiment classification module. 
The input review text $\mathbf{x}$ is first encoded by the shared text encoder into context-aware representations $\tilde{\mathbf{H}}=[\tilde{\mathbf{h}}_{1}, \ldots, \tilde{\mathbf{h}}_{L_{\mathbf{x}}}]$, which forms a memory bank for the summary decoder module and the dual-view sentiment classification module. 
Then, the summary decoder uses the memory bank from the encoder to compute a sequence of hidden states $\mathbf{S}=[\mathbf{s}_{1}, \ldots, \mathbf{s}_{L_{\mathbf{y}}}]$ and generates a review summary word by word. The ground-truth summary is used to compute a summary generation loss for the model. 
Our dual-view classification module consists of a source-view sentiment classifier and a summary-view sentiment classifier. The source-view sentiment classifier reads the encoder memory bank $\tilde{\mathbf{H}}$ and predicts a sentiment label $z$ for the review, while the summary-view sentiment classifier uses the decoder hidden states $\mathbf{S}$ to predict a sentiment label $z'$ for the generated summary. 
We use the ground-truth sentiment label of the review to compute a sentiment classification loss for both of these sentiment classifiers. 
Besides, we also introduce an inconsistency loss function to penalize the disagreement between these two classifiers. 
We jointly minimize all the above loss functions by a multi-task learning framework. 
During testing, we use the sentiment label predicted by the source-view classifier as the output sentiment label.
Figure~\ref{fig:architecture} illustrates the overall architecture of our model. We describe the details of each module as follows. 

\subsection{Shared Text Encoder (M1)}
The shared text encoder converts the review text into a memory bank for the sentiment classification and summary decoder modules. 
First, the encoder reads an input sequence $\mathbf{x}=(x_{1},\ldots,x_{L_{\mathbf{x}}})$ and uses a lookup table to convert each input word $x_{i}$ to a word embedding vector $\mathbf{x}_{i}\in\mathbb{R}^{d_{e}}$. 
To incorporate the contextual information of the review text into the representation of each word, we feed each embedding vector $\mathbf{x}_{i}$ to a bi-directional Gated-Recurrent Unit (GRU)~\cite{DBLP:conf/emnlp/ChoMGBBSB14} to learn a shallow hidden representation $\mathbf{u}_{i} \in \mathbb{R}^{d}$. 
More specifically, a bi-directional GRU consists of a forward GRU that reads the embedding sequence from $\mathbf{x}_{1}$ to $\mathbf{x}_{L_{\mathbf{x}}}$ and a backward GRU that reads from $\mathbf{x}_{L_{\mathbf{x}}}$ to $\mathbf{x}_{1}$: 
\begin{align}
    \overrightarrow{\mathbf{u}}_{i} &= \text{GRU}_{11}(\mathbf{x}_{i}, \overrightarrow{\mathbf{u}}_{i-1}) \text{,} \label{EQ:GRU11} \\
    \overleftarrow{\mathbf{u}}_{i} &= \text{GRU}_{12}(\mathbf{x}_{i}, \overleftarrow{\mathbf{u}}_{i+1}) \text{,}
\end{align}
where $\overrightarrow{\mathbf{u}}_{i}\in \mathbb{R}^{d/2}$ and $\overleftarrow{\mathbf{u}}_{i}\in \mathbb{R}^{d/2}$ denotes the hidden states of the forward GRU and backward GRU respectively. We concatenate $\overrightarrow{\mathbf{u}}_{i}$ and $\overleftarrow{\mathbf{u}}_{i}$ to form the shallow hidden representation for $\mathbf{x}_{i}$, i.e., $\mathbf{u}_{i}=[\overrightarrow{\mathbf{u}}_{i};\overleftarrow{\mathbf{u}}_{i}]$. 

Next, we pass the shallow hidden representations to another bi-directional GRU to model more complex interactions among the words in the review text: 
\begin{align}
    \overrightarrow{\mathbf{h}}_{i} &= \text{GRU}_{21}(\mathbf{u}_{i}, \overrightarrow{\mathbf{h}}_{t-1}) \text{,} \\
    \overleftarrow{\mathbf{h}}_{i} &= \text{GRU}_{22}(\mathbf{u}_{i}, \overleftarrow{\mathbf{h}}_{t+1}) \text{,}
\end{align}
where $\overrightarrow{\mathbf{h}}_{i}\in \mathbb{R}^{d/2}$ and $\overleftarrow{\mathbf{h}}_{i}\in \mathbb{R}^{d/2}$. 
We concatenate $\overrightarrow{\mathbf{h}}_{i}\in \mathbb{R}^{d/2}$ and $\overleftarrow{\mathbf{h}}_{i}$ to form $\mathbf{h}_{i} \in \mathbb{R}^{d}$. Then we apply a residual connection from the shallow hidden representation $\mathbf{u}_{i}$ to $\mathbf{h}_{i}$, which is standard technique to avoid gradient vanishing problem in deep neural networks~\cite{DBLP:conf/cvpr/HeZRS16}, as shown in the following equation:
\begin{align}
    \tilde{\mathbf{h}}_{i} &= \lambda \mathbf{h}_{i} + (1 - \lambda) \mathbf{u}_{i} \text{,}
\end{align}
where $\lambda \in [0,1]$ is a hyperparameter. We regard the final encoder representations $[\tilde{\mathbf{h}}_{1}, \ldots, \tilde{\mathbf{h}}_{L_{\mathbf{x}}}]$ as the memory bank for the later summary decoder and dual-view classification modules.

\subsection{Summary Decoder (M2)}
The decoder uses a forward GRU to generate an output summary $\mathbf{y}=[y_{1},\ldots,y_{t},\ldots,y_{L_{\mathbf{y}}}]$ step by step. The architecture of our summary decoder follows the decoder of the pointer generator network~\cite{DBLP:conf/acl/SeeLM17}. 
At each decoding step $t$, the forward GRU reads the embedding of the previous prediction $\mathbf{y}_{t-1}$ and the previous decoder hidden state $\mathbf{s}_{t-1}$ to yield the current decoder hidden state: 
\begin{align}
\mathbf{s}_{t}=\text{GRU}_{3}(\mathbf{y}_{t-1},\mathbf{s}_{t-1})\text{,}
\end{align}
where $\mathbf{s}_{t}\in \mathbb{R}^{d}$, $\mathbf{y}_{0}$ is the embedding of the start token. 
To gather relevant information from the document, a neural attention layer~\cite{DBLP:conf/iclr/BahdanauCB14} is then applied to compute an attention score $a_{t,i}$ between the current decoder hidden state $s_t$ and each of the vectors in the encoder memory bank $\tilde{\mathbf{h}}_{i}$:
\begin{align}
\alpha_{t,i} &= \mathbf{v}^{T}\tanh (\mathbf{W}_{h}\tilde{\mathbf{h}}_{i}+\mathbf{W}_{s}\mathbf{s}_{t}+\mathbf{b}_{attn})\text{, }\\
a_{t,i} &= \frac{\exp(\alpha_{t,i})}{\sum_{j=1}^{L_{\mathbf{x}}}\exp(\alpha_{t,j})}\text{,}
\end{align}
where $\mathbf{v}\in \mathbb{R}^{d'},\mathbf{W}_{h}\in \mathbb{R}^{d'\times d},\mathbf{W}_{s}\in \mathbb{R}^{d'\times d}\text{ and } \mathbf{b}_{attn}\in \mathbb{R}^{d'}$ are model parameters. The attention score is next used to compute a weighted sum of the memory bank vectors and produce an aggregated vector $\tilde{\mathbf{h}}^{*}_{t}$, which acts as the representation of the input sequence $\mathbf{x}$ at time $t$:
$\tilde{\mathbf{h}}^{*}_{t}=\sum_{i=1}^{L_{\mathbf{x}}} a_{t,i} \tilde{\mathbf{h}}_{i} \text{. }$
After that, we use $\tilde{\mathbf{h}}^{*}_{t}$ and the decoder hidden state $\mathbf{s}_{t}$ to compute a probability distribution over the words in a predefined vocabulary $\mathcal{V}$, as shown in the following equation, 
\begin{align}
P_{\mathcal{V}}(y_{t}|\mathbf{y}_{1:t-1},\mathbf{x})=\text{softmax}(\mathbf{W}_{V}(\mathbf{W}_{V'}[\mathbf{s}_{t};\tilde{\mathbf{h}}^{*}_{t}]+\mathbf{b}_{V'})+\mathbf{b}_{V})\text{, }
\end{align}
where $\mathbf{y}_{1:t-1}$ denotes the partial sequence of previous generated words, $[y_{1},\ldots,y_{t-1}]$, and $\mathbf{W}_{V}\in \mathbb{R}^{|\mathcal{V}|\times d},\mathbf{W}_{V'}\in \mathbb{R}^{d\times d},\mathbf{b}_{V}\in \mathbb{R}^{|\mathcal{V}|}, \mathbf{b}_{V'}\in \mathbb{R}^{d}$ are trainable parameters.

Although we can directly use $P_{\mathcal{V}}$ as our final prediction distribution, the decoder cannot generate out-of-vocabulary (OOV) words. To address this problem, we adopt the copy mechanism from~\cite{DBLP:conf/acl/SeeLM17} to empower the decoder to predict OOV words by directly copying words from the input review. In the copy mechanism, we first compute a soft gate $p_{gen}\in[0,1]$ between generating a word from the predefined vocabulary $\mathcal{V}$ according to $P_{\mathcal{V}}$ and copying a word from the source text $\mathbf{x}$ according to the attention distribution: 
\begin{align}
\label{eq:p-gen}
p_{gen}=\sigma(\mathbf{v}_{g}^{T} [\tilde{\mathbf{h}}^{*}_{t};\mathbf{s}_{t};\mathbf{y}_{t-1}]+\mathbf{b}_{gen})\text{, }
\end{align}
where $\mathbf{v}_{g}\in\mathbb{R}^{2d+d_{e}}$ and $\mathbf{b}_{gen}\in\mathbb{R}$ are trainable parameters and $\sigma(.)$ denotes the logistic sigmoid function. Next, we define a dynamic vocabulary $\mathcal{V}_{\mathbf{x}}$ as the union of $\mathcal{V}$ and all the words appear in the review $\mathbf{x}$. Finally, we can compute a probability distribution $P_{\mathcal{V}_{\mathbf{x}}}$ over the words in the dynamic vocabulary as follows:
\begin{align}
\label{eq:p-final}
P_{\mathcal{V}_{\mathbf{x}}}(y_{t})=p_{gen} P_{\mathcal{V}}(y_{t})+(1-p_{gen})\sum_{i:x_{i}=y_{t}} a_{t,i}\text{,}
\end{align}
where we use $P_{\mathcal{V}_{\mathbf{x}}}(y_{t})$ to denote $P_{\mathcal{V}_{\mathbf{x}}}(y_{t}|\mathbf{y}_{1:t-1},\mathbf{x})$ and $P_{\mathcal{V}}(y_{t})$ to denote $P_{\mathcal{V}}(y_{t}|\mathbf{y}_{1:t-1},\mathbf{x})$ for brevity. In Eq.~(\ref{eq:p-final}), we define $P_{\mathcal{V}}(y_{t})=0$ for all $y_{t}\notin \mathcal{V}$ (OOV words). If $y_{t}$ does not appear in $\mathbf{x}$, the copying probability, $\sum_{i:x_{i}=y_{t}} a_{t,i}$, will be zero.

\noindent \textbf{Summary generation loss function.} 
We use the negative log-likelihood of the ground-truth summary $\mathbf{y}^{*}$ as the loss function for the review summarization task: 
\begin{align}
\mathcal{L}_{gen}=-\sum_{t=1}^{L_{\mathbf{y}^{*}}} \log P_{\mathcal{V}_{\mathbf{x}}}(y^{*}_{t}|\mathbf{y}^{*}_{1:t-1},\mathbf{x})\text{,}
\end{align}
where $L_{\mathbf{y}^{*}}$ denotes the number of words in the ground-truth review summary $\mathbf{y}^{*}$. 

\noindent \textbf{Inference. }
In the testing stage, we use beam search to generate the output summary from the summary decoder. This is a standard technique to approximate the output sequence that achieves the highest generation probability.  

\subsection{Dual-view Sentiment Classification Module (M3)}\label{sec:dual-view-classifier}
We propose a dual-view sentiment classification module to learn a sentiment label for the review. It consists of a source-view sentiment classifier and a summary-view sentiment classifier. 

\noindent \textbf{Source-view sentiment classifier. }
The source-view sentiment classifier utilizes the encoder memory bank $\tilde{\mathbf{H}}=[\tilde{\mathbf{h}}_{1}, \ldots, \tilde{\mathbf{h}}_{L_{\mathbf{x}}}]$ to predict a sentiment label for the review. 
Since not all words in the review contribute equally to the prediction of sentiment label, we apply the attention mechanism~\cite{DBLP:conf/iclr/BahdanauCB14} to aggregate sentiment information from the encoder memory bank into a sentiment context vector. An additional glimpse operation~\cite{DBLP:journals/corr/VinyalsBK15} is incorporated in this aggregation process since the glimpse operation has been shown that it can improve the performance of several attention-based classification models~\cite{DBLP:journals/corr/BelloPLNB16,DBLP:conf/acl/BansalC18}.

First, a trainable query vector $q\in\mathbb{R}^{d_{q}}$ attends the encoder memory bank and produce a glimpse vector $\mathbf{g}\in\mathbb{R}^{d}$ as follows: 
\begin{align}
\label{eq:source-view-glimpse-1-1}
\alpha^{g}_{i} &= \mathbf{v}_{g}^{T} \tanh (\mathbf{W}_{gh} \tilde{\mathbf{h}}_{i} + \mathbf{W}_{gq} \mathbf{q} +\mathbf{b}_{g})\text{,}\\
a^{g}_{i} &= \frac{\exp(\alpha^{g}_{i})}{\sum_{j=1}^{L_{\mathbf{x}}}\exp(\alpha^{g}_{j})} \text{,} \\
\mathbf{g} &= \sum_{i=1}^{L_{\mathbf{x}}} a^{g}_{i} \tilde{\mathbf{h}}_{i} \text{,}
\end{align}
where $\mathbf{W}_{gh} \in \mathbb{R}^{d'\times d}$, $\mathbf{W}_{gq}\in \mathbb{R}^{d'\times d}$, $\mathbf{b}_{g}\in \mathbb{R}^{d'}$, $\mathbf{v}_{g}\in \mathbb{R}^{d'}$ are trainable model parameters. 

Then, the glimpse vector $\mathbf{g}$ attends the memory bank again to compute a review sentiment context vector $\mathbf{e}\in\mathbb{R}^{d}$: $\mathbf{e} = \sum_{i=1}^{L_{\mathbf{x}}} a^{e}_{i} \tilde{\mathbf{h}}_{i} \text{,}$ where $a^{e}_{i}$ is the corresponding attention weight.

The classifier is a two-layer feed-forward neural network using a rectified linear unit (ReLU) as the activation function~\cite{DBLP:conf/ijcai/MaSLR18}. The softmax function is applied at the output of the network to produce a probability distribution over the sentiment label: 
\begin{align}
    P_{ec}(z|\mathbf{x}) = \text{softmax}(\mathbf{W}_{z2}(\text{ReLU}(\mathbf{W}_{z1} \mathbf{e} + \mathbf{b}_{z1}))+\mathbf{b}_{z2})\text{,}
\end{align}
where $\mathbf{W}_{z1}\in \mathbb{R}^{d_{z}\times d}$, $\mathbf{W}_{z2}\in \mathbb{R}^{K\times d_{z}}$, $\mathbf{b}_{z1}\in \mathbb{R}^{d_{z}}$, $\mathbf{b}_{z2}\in \mathbb{R}^{K}$ are model parameters. 
The sentiment label with the highest probability is treated as the predicted sentiment label for the review. 

\noindent \textbf{Summary-view sentiment classifier. }
The summary-view sentiment classifier uses the decoder hidden states $\mathbf{S}=[\mathbf{s}_{1}, \ldots, \mathbf{s}_{L_{\mathbf{y}}}]$ to predict a sentiment label for the generated summary. 
Since each decoder hidden state is used to generate a summary word, we treat the decoder hidden states as the representation for the generated summary. Then, we apply attention mechanism~\cite{DBLP:conf/iclr/BahdanauCB14} with glimpse operation~\cite{DBLP:journals/corr/VinyalsBK15} to compute a sentiment context vector $\mathbf{e}'$ for the generated summary. The architecture of the summary-view sentiment classifier is the same as the source-view sentiment classifier but with another set of parameters. The only difference is that it uses the decoder hidden states as the input instead of the encoder memory bank, i.e., all the $\tilde{\mathbf{h}}_{i}$ terms in the equations of the source-view sentiment classifier are replaced by $\mathbf{s}_{i}$. Then, the summary-view sentiment classifier outputs a probability distribution over the sentiment label for the generated summary: $P_{dc}(z'|\mathbf{x},\mathbf{y})$.

\noindent \textbf{Sentiment classification loss function. }
We use negative log-likelihood as the classification loss function for both the source-view sentiment classifier and the summary-view sentiment classifier:  
\begin{align}
    \mathcal{L}_{ec} &= - \log P_{ec}(z^{*}|\mathbf{x}) \text{,} \\
    \mathcal{L}_{dc} &= - \log P_{dc}(z^*|\mathbf{x},\mathbf{y}) \text{,}
\end{align}
where $z^*$ denotes the ground-truth sentiment label, $\mathcal{L}_{ec}$ and $\mathcal{L}_{dc}$ denote the classification loss for the source-view and the summary-view sentiment classifiers respectively. 

\begin{table*}[t]
\caption{Statistics of the datasets. ``\textit{Ave. RL}'' denotes the average review length (in words) in the training set. ``\textit{Ave. SL}'' denotes the average summary length in the training set. ``\textit{Copy Ratio}'' indicates the ratio of the summary words that are copied from the review in the training set. ``\textit{Rating k}'' means the ratio of the data samples with sentiment label \textit{k} in the training set. }
\label{tab:datasets}
\begin{tabular}{|l| c c c| c c| c| c c c c c|}
\hline 
\textbf{Dataset} & \textit{Training} & \textit{Validation} & \textit{Testing}  &  \textit{Ave. RL} & \textit{Ave. SL} & \textit{Copy Ratio} & \textit{Rating 1} & \textit{Rating 2} & \textit{Rating 3} & \textit{Rating 4} & \textit{Rating 5} \\
\hline 
\hline
 \textbf{Sports} & 183,714 & 9,000 & 9,000 & 108.3 & 6.7  & 58.5\% & 3.3\% & 3.9\% & 9.1\% & 22.9\% & 60.8\% \\
\hline
 \textbf{Movies} & 1,200,601 & 20,000 & 20,000 & 167.1 & 6.6 & 58.8\% & 6.3\% & 6.3\% & 12.6\% & 23.3\% & 51.5\% \\
\hline
\textbf{Toys} & 104,296 & 8,000 & 8,000 & 125.9 & 6.8 & 61.0\% & 2.9\% & 4.1\% & 11.0\% & 24.0\% & 58.0\% \\
\hline
\textbf{Home} & 367,395 & 10,000 & 10,000 & 120.9 & 6.8 & 59.8\% & 5.3\% & 4.9\% & 9.1\% & 20.1\% & 60.6\% \\
\hline
\end{tabular}
\end{table*}

\noindent \textbf{Inconsistency loss function. }
We introduce an inconsistency loss to penalize the disagreement between the source-view sentiment classifier and the summary-view sentiment classifier. The intuition is that the review summary should have the same sentiment attitude as the input review. We define our inconsistency loss function as the Kulllback-Leibler (KL) divergence between $P_{ec}$ and $P_{dc}$: 
\begin{align}
    \mathcal{L}_{inc} = D_{KL}(P_{ec}||P_{dc}) = \sum_{k=1}^{K} P_{ec}(k|\mathbf{x}) \log \frac{P_{ec}(k|\mathbf{x})}{P_{dc}(k|\mathbf{x},\mathbf{y})} \text{.}
\end{align}
Since the summary-view sentiment classifier uses the decoder hidden states to predict the sentiment label for the generated summary, the inconsistency loss encourages the sentiment information in the decoder states to be close to that in the encoder memory bank. Thus, it helps the decoder to generate a summary that has a consistent sentiment with the review. 
Moreover, the source-view and summary-view sentiment classifiers can learn from each other to improve the sentiment classification performance. 

\noindent \textbf{Inference.}
In the testing stage, we use the sentiment label predicted by the source-view classifier as the final classification prediction. The reason is that the decoder suffers from the well-known exposure bias problem~\cite{DBLP:journals/corr/RanzatoCAZ15} during testing, which affects the performance of the summary-view classifier when inference. We conduct experiments to analyze the influence of the exposure bias on the classification performance and provide more discussions in Section~\ref{sec:classifier-exp}. 

\subsection{Multi-task Training Objective}
We adopt a multi-task learning framework to jointly minimize the review summarization loss, source-view sentiment classification loss, summary-view sentiment classification loss, and inconsistency loss. The objective function is shown as follows. 
\begin{align}
    \label{eq:training-objective}
    \mathcal{L} = \gamma_{1} \mathcal{L}_{gen} + \gamma_{2} \mathcal{L}_{ec} + \gamma_{3} \mathcal{L}_{dc} + \gamma_{4} \mathcal{L}_{inc} \text{,}
\end{align}
where $\gamma_{1},\gamma_{2},\gamma_{3},\gamma_{4}$ are hyper-parameters that controls the weights of these four loss functions. We set $\gamma_{1}=0.8,\gamma_{2}=0.1,\gamma_{3}=0.1,\gamma_{4}=0.1$ after fine-tuning on the validation datasets. Thus, each module of our joint model can be trained end-to-end. 

\section{Experimental Setup}

\subsection{Datasets}
In this work, we conduct experiments on four real-world datasets from different domains. The datasets are collected from the Amazon 5-core review repository~\cite{DBLP:conf/sigir/McAuleyTSH15}. We adopt product reviews from the following four domains as our datasets: \textbf{Sports \& Outdoors}; \textbf{Movies \& TV}; \textbf{Toys \& Games}; and \textbf{Home \& Kitchen}. 
In our experiments, each data sample consists of a review text, a summary, and a rating. We regard the rating as a sentiment label, which is an integer in the range of $[1,5]$. For text preprocessing, we lowercase all the letters and tokenize the text using Stanford CoreNLP~\cite{DBLP:conf/acl/ManningSBFBM14}. We append a period to a summary sentence if it is not ended properly. To reduce the noise in these datasets, we filter out data samples when the review length is less than 16 or longer than 800, or the summary length is less than 4. 
We randomly split each dataset into training, validation, and testing sets. The statistics of these datasets are shown in Table~\ref{tab:datasets}. 


\subsection{Evaluation Metrics}
For review summarization, we use \textbf{ROUGE score}~\cite{lin2004rouge} as the evaluation metric, which is a standard evaluation metric in the field of text summarization\cite{DBLP:conf/acl/SeeLM17, DBLP:conf/sigir/LiWRBL17}. 
Following \cite{DBLP:conf/ijcai/MaSLR18}, we use ROUGE-1, ROUGE-2, and ROUGE-L scores to evaluate the qualities of the generated review summaries. 
ROUGE-1 and ROUGE-2 measure the overlapping uni-grams and bi-grams between the generated review summary $\mathbf{y}$ and the ground-truth review summary $\mathbf{y}^{*}$. 
ROUGE-L measures the longest common subsequence between the generated summary and the ground-truth review summary, we refer the readers to \cite{lin2004rouge} for details. 

For sentiment classification, we use the \textbf{macro $F_{1}$ score} and the \textbf{balanced accuracy}~\cite{DBLP:conf/icpr/BrodersenOSB10} as the evaluation metrics. We denote the macro $F_{1}$ score as ``M. $F_{1}$'' and the balanced accuracy as ``B. Acc''. From Table~\ref{tab:datasets}, we can observe that the class distribution of the sentiment labels is imbalanced. Therefore, we do not use overall accuracy as the evaluation metric. 
To compute macro $F_{1}$ score, we first calculate the precision $p_{i}$ and recall $r_{i}$ for each individual sentiment class $i$. 
Next, we compute the macro-averaged precision and recall as follows, $p_{macro}=\sum_{i=1}^{K}p_{i}/K$, $r_{macro} = \sum_{i=1}^{K}r_{i}/K$. 
The macro $F_{1}$ score is the harmonic mean of $p_{macro}$ and $r_{macro}$. 
The balanced accuracy is a variant of the accuracy metric for imbalanced datasets~\cite{DBLP:conf/icpr/BrodersenOSB10,kelleher2015fundamentals}. It is defined as the macro-average of the recall obtained on each class, i.e., $\text{balanced accuracy}=r_{macro}$.

\subsection{Baselines}
Our baselines are categorized into three groups: (1) summarization-only models; (2) sentiment-only models; and (3) joint models.
We use the following summarization-only models as our review summarization baselines.
\begin{itemize}
    \item \textbf{PGNet}~\cite{DBLP:conf/acl/SeeLM17}: A popular summarization model which is based on the encoder-decoder framework with attention and copy mechanisms. 
    \item \textbf{C.Transformer}~\cite{DBLP:conf/emnlp/GehrmannDR18}: The CopyTransformer model that enhances the state-of-the-art Transformer model~\cite{DBLP:conf/nips/VaswaniSPUJGKP17} with copy mechanism~\cite{DBLP:conf/acl/SeeLM17} for abstractive summarization. 
\end{itemize}
The following sentiment-only model are employed as one of our sentiment classification baselines. 
\begin{itemize}
    \item \textbf{BiGRU+Attention}: A bi-directional GRU layer~\cite{DBLP:conf/emnlp/ChoMGBBSB14} first encodes the input review into hidden states. Then it uses an attention mechanism~\cite{DBLP:conf/iclr/BahdanauCB14} with glimpse operation~\cite{DBLP:journals/corr/VinyalsBK15} to aggregate information from the encoder hidden states and produce a vector. The vector is then fed through a two-layer feed-forward neural network to predict the sentiment label. 
    \item \textbf{DARLM}:~\cite{DBLP:conf/ijcai/ZhouWD18} The state-of-the-art model for sentence classification. 
\end{itemize}
We also use the following joint models as the baselines of both review summarization and sentiment classification. 
\begin{itemize}
   \item \textbf{HSSC}~\cite{DBLP:conf/ijcai/MaSLR18}: The state-of-the-art model for jointly improving review summarization and sentiment classification. 
   \item \textbf{Max}~\cite{DBLP:conf/ijcai/MaSLR18}: A bi-directional GRU layer first encodes the input review into hidden states. These hidden states are then shared by a summary decoder and a sentiment classifier. The sentiment classifier uses max pooling to aggregate the the encoder hidden states into a vector, which is then fed through a two-layer feed-forward neural network to predict the sentiment label. 
   \item \textbf{HSSC+copy}: We incorporate the copy mechanism~\cite{DBLP:conf/acl/SeeLM17} into the HSSC model~\cite{DBLP:conf/ijcai/MaSLR18} as a strong baseline.
   \item \textbf{Max+copy}: We also incorporate the copy mechanism~\cite{DBLP:conf/acl/SeeLM17} into the Max model as another strong baseline.
\end{itemize}

\subsection{Implementation Details}
We train a word2vec~\cite{DBLP:conf/nips/MikolovSCCD13} with a dimension of 128 (i.e., $d_{e}=128$) on the training set of each dataset to initialize the word embeddings of all the models including baselines. All the initialized embeddings are fine-tuned by back-propagation during training. The vocabulary $\mathcal{V}$ is defined as the 50,000 most frequent words in the training set. 
The hidden sizes $d$, $d'$, $d_{q}$, and $d_{z}$ are set to 512. The $\lambda$ in the residual connection is set to 0.5. The initial hidden states of the GRU cells in shared text encoder are zero vectors, while the initial hidden states of the GRU cell in the summary decoder is set to $[\lambda \overrightarrow{\mathbf{h}}_{L_{\mathbf{x}}} + (1-\lambda) \overrightarrow{\mathbf{u}}_{L_{\mathbf{x}}} ; \lambda \overleftarrow{\mathbf{h}}_{1} + (1-\lambda) \overleftarrow{\mathbf{u}}_{1}]$. A dropout layer with $p=0.1$ is applied at the two-layer feed-forward neural networks in source-view and summary-view sentiment classifiers. 
The hidden sizes of the GRU cells and the feed-forward neural networks in our baselines are set to 512, which is the same as our model. 
During training, we truncate the input review text to 400 tokens and the output summary to 100 tokens. We use the Adam optimization algorithm~\cite{DBLP:journals/corr/KingmaB14} with a batch size of 32 and an initial learning rate of 0.001. We use the gradient clipping of 2.0 using L-2 norm. The learning rate is reduced by half if the validation loss stops dropping. We apply early stopping when the validation loss stops decreasing for three consecutive checkpoints. 
During testing, we use the beam search algorithm with a beam size of 5 and a maximum depth of 120 for all models. 
We repeat all experiments five times with different random seeds and the averaged results are reported\footnote{Source code is available at https://github.com/kenchan0226/dual\_view\_review\_sum}.

\begin{table*}[t]
\centering
\caption{Review summarization results on the four datasets. We use `R' to denote recall and `P' to denote precision. 
}
\label{table:summary results}
\resizebox{\textwidth}{!}{
    \begin{tabular}{|c | c | c | c | c | c | c | c | c | c | c |}
        \hline
        \multirow{2}{*}{\textbf{Dataset}} & 
        \multirow{2}{*}{\textbf{Method}} & 
        \multicolumn{3}{c|}{\textbf{ROUGE-1}} & 
        \multicolumn{3}{c|}{\textbf{ROUGE-2}} & 
        \multicolumn{3}{c|}{\textbf{ROUGE-L}} \\ \cline{3-11}
        &  & R  & P & $F_{1}$ & R  & P & $F_{1}$ & R  & P & $F_{1}$ \\
        \hline
        \multirow{6}{*}{Sports}
        & PGNet & 14.78$\pm$.21 &  19.79$\pm$.22 & 16.13$\pm$.21 & 6.13$\pm$.14 & 8.20$\pm$.16 & 6.62$\pm$.15 & 14.58$\pm$.20 & 19.46$\pm$.22 & 15.89$\pm$.21\\
        \cline{2-11}
        & C.Transformer & 13.73$\pm$.11 & 18.46$\pm$.12 & 15.02$\pm$.11 & 5.13$\pm$.12 & 6.88$\pm$.16 & 5.56$\pm$.13 & 13.54$\pm$.11 & 18.13$\pm$.12 & 14.80$\pm$.11 \\
        \cline{2-11}
        & HSSC  & 14.44$\pm$.25 & 19.24$\pm$.30 & 15.74$\pm$.26 & 5.52$\pm$.13 & 7.24$\pm$.16 & 5.93$\pm$.13 & 14.24$\pm$.24 & 18.91$\pm$.30 & 15.50$\pm$.25 \\
         \cline{2-11}
              & Max   & 14.36$\pm$.29 & 19.20$\pm$.30 & 15.67$\pm$.27 & 5.46$\pm$.14 & 7.22$\pm$.17 & 5.88$\pm$.13 & 14.14$\pm$.28 & 18.85$\pm$.29 & 15.41$\pm$.26 \\
         \cline{2-11}
          & HSSC+copy & 14.64$\pm$.29 & 19.61$\pm$.41 & 15.98$\pm$.32 & 5.95$\pm$.13 & 7.98$\pm$.12 & 6.43$\pm$.12 & 14.43$\pm$.28 & 19.26$\pm$.39 & 15.74$\pm$.31 \\
         \cline{2-11}
          & Max+copy  & 14.75$\pm$.29 & 19.86$\pm$.37 & 16.15$\pm$.32 & 6.11$\pm$.17 & 8.22$\pm$.23 & 6.62$\pm$.18 & 14.56$\pm$.27 & 19.53$\pm$.35 & 15.92$\pm$.30 \\
          \cline{2-11}
          & \textbf{Dual-view} & \textbf{15.39$\pm$.22} & \textbf{20.53$\pm$.32} & \textbf{16.79$\pm$.26} & \textbf{6.46$\pm$.07} & \textbf{8.63$\pm$.13} & \textbf{6.98$\pm$.10} & \textbf{15.18$\pm$.22} & \textbf{20.19$\pm$.31} & \textbf{16.55$\pm$.25} \\
        \hline
        \hline
      \multirow{6}{*}{Movie}
        & PGNet & 12.67$\pm$.05 & 17.76$\pm$.06 & 14.04$\pm$.05 & 5.14$\pm$.07 & 7.38$\pm$.06 & 5.66$\pm$.06 & 12.40$\pm$.4 & 17.32$\pm$.05 & 13.72$\pm$.4\\
        \cline{2-11}
        & C.Transformer  & 12.09$\pm$.14 & 16.78$\pm$.14 & 13.34$\pm$.13 & 4.46$\pm$.05 & 6.30$\pm$.07 & 4.89$\pm$.05 & 11.81$\pm$.14 & 16.33$\pm$.14 & 13.01$\pm$.13 \\
          \cline{2-11}
        & HSSC & 12.59$\pm$.06 & 17.62$\pm$.09 & 13.89$\pm$.07 & 4.62$\pm$.05 & 6.59$\pm$.07 & 5.06$\pm$.05 & 12.27$\pm$.06 & 17.12$\pm$.09 & 13.53$\pm$.08 \\
        \cline{2-11}
        & Max  & 12.49$\pm$.10 & 17.53$\pm$.21 & 13.82$\pm$.14 & 4.70$\pm$.16 & 6.72$\pm$.34 & 5.16$\pm$.21 & 12.20$\pm$.10 & 17.07$\pm$.22 & 13.48$\pm$.15 \\
        \cline{2-11}
            & HSSC+copy & 12.66$\pm$.06 & 17.92$\pm$.12 & 14.08$\pm$.05 & 5.06$\pm$.07 & 7.37$\pm$.09 & 5.60$\pm$.07 & 12.39$\pm$.06 & 17.47$\pm$.14 & 13.76$\pm$.07 \\
        \cline{2-11}
            & Max+copy  & 12.61$\pm$.04 & 17.81$\pm$.11 & 14.01$\pm$.06 & 5.04$\pm$.05 & 7.32$\pm$.12 & 5.57$\pm$.07 & 12.34$\pm$.04 & 17.38$\pm$.09 & 13.69$\pm$.05 \\
        \cline{2-11}
        
          & \textbf{Dual-view}  & \textbf{12.84$\pm$.07} & \textbf{17.98$\pm$.04} & \textbf{14.22$\pm$.06} & \textbf{5.22$\pm$.05} & \textbf{7.48$\pm$.06} & \textbf{5.75$\pm$.05} & \textbf{12.57$\pm$.07} & \textbf{17.55$\pm$.07} & \textbf{13.90$\pm$.06} \\
        \hline
        \hline
        \multirow{6}{*}{Toys}
        & PGNet  & 14.77$\pm$.24 & 20.54$\pm$.43 & 16.40$\pm$.27 & 6.18$\pm$.18 & 8.47$\pm$.19 & 6.74$\pm$.17 & 14.53$\pm$.23 & 20.13$\pm$.40 & 16.11$\pm$.25 \\
        \cline{2-11}
        & C.Transformer  & 12.57$\pm$.31 & 17.53$\pm$.53 & 13.94$\pm$.35 & 4.76$\pm$.08 & 6.42$\pm$.19 & 5.14$\pm$.10 & 12.36$\pm$.29 & 17.16$\pm$.51 & 13.69$\pm$.32 \\
        \cline{2-11}
        & HSSC & 14.04$\pm$.20 & 19.29$\pm$.45 & 15.48$\pm$.26 & 5.24$\pm$.11 & 7.03$\pm$.16 & 5.66$\pm$.10 & 13.76$\pm$.19 & 18.81$\pm$.42 & 15.16$\pm$.24 \\
        \cline{2-11}
        & Max  & 14.15$\pm$.26 & 19.36$\pm$.24 & 15.59$\pm$.23 & 5.27$\pm$.38 & 6.99$\pm$.40 & 5.66$\pm$.37 & 13.90$\pm$.26 & 18.92$\pm$.25 & 15.29$\pm$.24 \\
        \cline{2-11}
        & HSSC+copy & 14.70$\pm$.23 & 20.29$\pm$.33 & 16.27$\pm$.25 & 6.18$\pm$.14 & 8.38$\pm$.24 & 6.71$\pm$.17 & 14.46$\pm$.23 & 19.88$\pm$.30 & 15.98$\pm$.25 \\
        \cline{2-11}
        & Max+copy & 14.62$\pm$.22 & 20.52$\pm$.59 & 16.29$\pm$.31 & 5.92$\pm$.16 & 8.19$\pm$.15 & 6.48$\pm$.11 & 14.37$\pm$.19 & 20.09$\pm$.59 & 15.99$\pm$.29 \\
      \cline{2-11}
        & \textbf{Dual-view} & \textbf{14.83$\pm$.30} & \textbf{20.76$\pm$.41} & \textbf{16.50$\pm$.30} & 6.17$\pm$.31 & \textbf{8.57$\pm$.23} & \textbf{6.75$\pm$.26} & \textbf{14.57$\pm$.28} & \textbf{20.30$\pm$.36} & \textbf{16.19$\pm$.26} \\
        \hline
        \hline
        \multirow{6}{*}{Home}
        & PGNet & 14.82$\pm$.19 & 20.53$\pm$.22 & 16.44$\pm$.19 & 6.28$\pm$.13 & 8.83$\pm$.15 & 6.90$\pm$.13 & 14.64$\pm$.18 & 20.23$\pm$.21 & 16.23$\pm$.18 \\
        \cline{2-11}
        & C.Transformer  & 13.75$\pm$.13 & 19.35$\pm$.15 & 15.36$\pm$.13 & 5.44$\pm$.13 & 7.70$\pm$.14 & 6.01$\pm$.13 & 13.58$\pm$.12 & 19.06$\pm$.15 & 15.17$\pm$.13 \\
        \cline{2-11}
        & HSSC & 14.71$\pm$.17 & 20.29$\pm$.26 & 16.28$\pm$.20 & 5.85$\pm$.14 & 8.10$\pm$.19 & 6.39$\pm$.15 & 14.51$\pm$.17 & 19.97$\pm$.26 & 16.05$\pm$.20 \\
        \cline{2-11}
        & Max & 14.85$\pm$.13 & 20.35$\pm$.14 & 16.39$\pm$.13 & 5.97$\pm$.08 & 8.26$\pm$.11 & 6.53$\pm$.09 & 14.65$\pm$.11 & 20.03$\pm$.12 & 16.15$\pm$.11 \\
        \cline{2-11}
        & HSSC+copy & 14.93$\pm$.17 & 20.62$\pm$.17 & 16.54$\pm$.17 & 6.34$\pm$.11 & 8.87$\pm$.14 & 6.95$\pm$.11 & 14.75$\pm$.17 & 20.32$\pm$.17 & 16.33$\pm$.17 \\
        \cline{2-11}
        & Max+copy & 14.92$\pm$.13 & 20.57$\pm$.17 & 16.52$\pm$.14 & 6.33$\pm$.11 & 8.84$\pm$.14 & 6.94$\pm$.11 & 14.72$\pm$.12 & 20.25$\pm$.17 & 16.30$\pm$.13 \\
      \cline{2-11}
      & \textbf{Dual-view} & \textbf{15.18$\pm$.11} & \textbf{20.96$\pm$.09} & \textbf{16.81$\pm$.09} & \textbf{6.57$\pm$.11} & \textbf{9.19$\pm$.14} & \textbf{7.21$\pm$.12} & \textbf{15.00$\pm$.10} & \textbf{20.65$\pm$.07} & \textbf{16.60$\pm$.08} \\
        \hline
    \end{tabular}
}
\end{table*}

\begin{table*}[t]
\centering
\caption{Sentiment classification results on the four datasets. 
}
\label{table:sentiment results}
\begin{tabular}{| c | c  c | c  c | c  c | c  c |}
        \hline
        \multirow{2}{*}{\textbf{Method}} & 
        \multicolumn{2}{c|}{\textbf{Sports}} & 
        \multicolumn{2}{c|}{\textbf{Movies}} & 
        \multicolumn{2}{c|}{\textbf{Toys}} &
        \multicolumn{2}{c|}{\textbf{Home}} \\  \cline{2-9}
        & M. $F_{1}$ & B. Acc & M. $F_{1}$ & B. Acc & M. $F_{1}$ & B. Acc & M. $F_{1}$ & B. Acc \\
        \hline
        HSSC              & 53.49$\pm$.83  & 51.99$\pm$1.79  & 60.67$\pm$.40 & 59.23$\pm$.55 & 54.24$\pm$1.05 & 53.66$\pm$.97 & 58.51$\pm$.46 & 57.42$\pm$.83 \\
        Max               & 53.27$\pm$1.70 & 52.64$\pm$1.98 & 60.66$\pm$.39 & 59.34$\pm$.53 & 55.02$\pm$.49 & 53.64$\pm$.68 & 58.31$\pm$.78 & 57.36$\pm$.86 \\
        \hline
        BiGRU+Attention   & 54.21$\pm$1.21 & 53.03$\pm$2.55 & 61.14$\pm$.38 & 59.80$\pm$.47 & 53.54$\pm$2.84 & 52.82$\pm$2.94 & 59.32$\pm$.72 & 58.03$\pm$.55 \\
        DARLM             & 49.60$\pm$1.85 & 47.95$\pm$.83  & 57.75$\pm$.43 & 53.96$\pm$.83 & 50.58$\pm$.45 & 48.67$\pm$.98 & 54.49$\pm$1.82 & 53.43$\pm$.83 \\
        \hline
        HSSC+copy         & 53.14$\pm$1.33 & 52.63$\pm$.45  & 60.68$\pm$.37 & 59.32$\pm$.50 & 54.38$\pm$.86 & 53.32$\pm$1.26 & 58.78$\pm$.80 & 58.02$\pm$.70 \\
        Max+copy          & 53.95$\pm$1.19 & 52.53$\pm$.36  & 60.60$\pm$.36 & 59.25$\pm$.59 & 53.52$\pm$1.60 & 52.01$\pm$1.43 & 58.85$\pm$.41 & 58.05$\pm$.46 \\
        \hline
         \textbf{Dual-view}   & \textbf{56.31$\pm$.67} & \textbf{54.28$\pm$.63} 
                              & \textbf{62.00$\pm$.19} & \textbf{60.52$\pm$.21}
                              & \textbf{55.70$\pm$.86} & \textbf{54.06$\pm$1.57} 
                              & \textbf{60.73$\pm$.55} & \textbf{59.63$\pm$.52} \\
        \hline
\end{tabular}
\end{table*}

\section{Results Analysis}
Our experiments are intended to address the following research questions. 
\begin{itemize}
    \item \textbf{RQ1:} What is the performance of our proposed model on review summarization and sentiment classification?
    \item \textbf{RQ2:} What is the impact of each component of our model on the overall performance? 
    \item \textbf{RQ3:} Which of the source-view and summary-view classifiers is better? Can we further improve the sentiment classification performance if we combine the source-view and summary-view classifiers by ensemble?
    \item \textbf{RQ4:} Are the generated review summaries consistent with the predicted sentiment labels?
\end{itemize}

\subsection{Main Results (RQ1)}
We show the review summarization results on the four datasets in Table~\ref{table:summary results}. We note that our dual-view model achieves the best performance on almost all the metrics among all the four real-world datasets, demonstrating the effectiveness of our model to summarize a product review on different domains. 
We also conduct a significance test comparing our model with HSSC, Max, and PGNet. The results show our dual-view model significantly outperforms these three baselines on most of the metrics (paired t-test, $p<0.05$). 

The review sentiment classification results are shown in Table~\ref{table:sentiment results}. For our dual-view model, the classification results of the source-view classifier are reported in this table. We observe that our dual-view model consistently outperforms all the baseline methods on all the datasets. These results show that our model can predict more accurate sentiment labels than baselines. Besides, the significance test results comparing with HSSC and Max indicate the improvements by our dual-view model are significant on most of the metrics (paired t-test, $p<0.05$). 


\begin{table}[t]
\centering
\caption{Ablation study on the Sports dataset. ``Full'' indicates our full model. ``-I'' means the inconsistency loss is removed. ``-A'' means we replace the attention based classifier with a max pooling based classifier. ``-R'' means we remove the residual connection in the encoder part and only use one GRU layer as the encoder. ``-C'' indicates the copy mechanism is removed. RG-* is the $F_1$-Measure of ROUGE-*. 
}
\label{table:ablation-study}
\begin{tabular}{|l | c c c | c | c|}
     \hline
     & RG-1 & RG-2 & RG-L & M. $F_1$ & B. Acc \\
     \hline
     Full                    &  \textbf{16.79} & \textbf{6.98} & \textbf{16.55} & \textbf{56.31} & \textbf{54.28}\\
     -I                       &  16.38 & 6.70 & 16.15 & 55.95 & 53.95\\
     -A                       &  16.50 & 6.81 & 16.25 & 55.36 & 53.97\\
     -R                       &  16.31 & 6.64 & 16.06 & 54.38 & 51.99\\
     -C                       &  15.33 & 5.72 & 15.10 & 55.92 & 54.27\\
     \hline
\end{tabular}
\end{table}

\subsection{Ablation Study (RQ2)}
We conduct an ablation study to verify the effectiveness of each important component of our model. The results on the Sports dataset are displayed in Table~\ref{table:ablation-study}. 
``-I'' denotes that we do not incorporate the inconsistency loss when training. By comparing the performance of the full model and ``-I'' in the table, we observe that after removing the inconsistency loss, the performance of both review summarization and sentiment classification drops obviously. Our experiment results on the Sports validation set also show that our inconsistency loss substantially reduces the number of inconsistent predicted sentiment labels between the source-view and summary-view sentiment classifiers from 11.9\% to 6.3\%. 
If we replace the attention mechanism in classifiers with a max-pooling operation (i.e., compare ``Full'' and ``-A'' in the table), the performance decreases as we anticipated.
We also find that after removing the residual connection in the encoder (i.e., compare ``Full'' and ``-R''), the performance of both review summarization and sentiment classification degrades, which suggests that the residual connection module is effective for both tasks. Moreover, we note that the copy mechanism (i.e., compare ``Full'' and ``-C'') is helpful for both review summarization and sentiment classification.

\begin{table}[t]
\centering
\caption{The macro $F_{1}$ scores of the source-view, summary-view, and merged sentiment classifiers on the four datasets. 
}
\label{table:sentiment results of three classifiers}
\begin{tabular}{| c | c  c c  c |}
        \hline
        \textbf{Classifiers} & \textbf{Sports} & \textbf{Movies} & \textbf{Toys} & \textbf{Home} \\
        \hline
         Source-view    & \textbf{56.31} & \textbf{62.00} & \textbf{55.70} & 60.73 \\
         Summary-view  & 55.56 & 61.55 & 55.02 & 60.48 \\
         Merged        & 56.28 & 61.90 & 55.55  & \textbf{60.92} \\
        \hline
\end{tabular}
\end{table}

\begin{table}[t]
\centering
\caption{The macro $F_{1}$ scores of the source-view, summary-view, and merged sentiment classifiers with and without teacher-forcing on Sports validation and testing datasets. ``w/ TF'' means we apply teacher forcing when generating summaries. ``w/o TF'' indicates teacher forcing is removed. 
}
\label{table:sentiment results with or without TF}
\begin{tabular}{| c | c  c | c  c |}
        \hline
        \multirow{2}{*}{\textbf{Classifiers}} &  \multicolumn{2}{c|}{\textbf{Validation Set}} & \multicolumn{2}{c|}{\textbf{Testing Set}} \\
        \cline{2-5}
        & w/ TF & w/o TF & w/TF & w/o TF \\
        \hline
         Source-view   & 54.81 & 54.81 & 56.31 & 56.31 \\
         Summary-view  & 56.86 & 54.26 & 58.62 & 55.56 \\
         Merged        & 56.06 & 54.86 & 57.44 & 56.28 \\
        \hline
\end{tabular}
\end{table}

\subsection{Classifier Ensemble (RQ3)}\label{sec:classifier-exp}
Though Table~\ref{table:sentiment results} reports the performance of the source-view sentiment classifier of our dual-view model, we are also interested in the performance of the summary-view sentiment classifier of our model and whether merging these two classifiers can improve the sentiment classification performance. 
To combine the sentiment label predictions from the source-view and summary-view sentiment classifiers, we average their predicted sentiment label probability distributions into a merged prediction probability distribution: $p_{mrg} = \frac{P_{ec} + P_{dc}}{2}$.
We report the macro $F_{1}$ scores of the source-view, summary view, and merged sentiment classifiers in Table~\ref{table:sentiment results of three classifiers}. 

From this table, we note that the source-view classifier achieves the best results on three datasets. Thus, combining the source-view and summary-view sentiment classifiers does not yield better performance in most of the cases. Moreover, we also find that the source-view sentiment classifier consistently outperforms the summary-view sentiment classifier on all of these datasets. 
The main reason is that the exposure bias issue~\cite{DBLP:journals/corr/RanzatoCAZ15} of RNN decoder affects the performance of the summary-view sentiment classifier during testing. 
More specifically, in the training stage, previous ground-truth summary tokens are fed into the decoder to predict the next summary token (i.e., teacher-forcing) and the hidden states of the decoder can be regarded as the hidden representations of the ground-truth summary. 
In the testing stage, we cannot access the ground-truth summary. The previously-predicted tokens are fed into the decoder to predict the next summary token and the errors are accumulated in the decoder hidden states. Therefore, the summary-view classifier, which is based on the decoder hidden states, has a worse performance during testing. 

We conduct experiments to demonstrate the effects of the exposure bias problem. Table~\ref{table:sentiment results with or without TF} displays the macro $F_{1}$ scores of different sentiment classifiers with and without teacher-forcing. 
It is observed that when teacher-forcing is applied (i.e., w/ TF in the table), the summary-view classifier outperforms the source-view classifier by a large margin. However, its performance drops dramatically on both validation and testing sets after removing teacher-forcing (i.e., w/o TF). As we anticipated, the source-view classifier is not affected by whether teacher-forcing is adopted since it performs sentiment classification from the source input view instead of the summary view. 
The results of balanced accuracy scores show similar observations and we do not report them in Table~\ref{table:sentiment results of three classifiers} and~\ref{table:sentiment results with or without TF} for brevity. 
These results suggest a future work of alleviating the performance gap of the summary-view sentiment classifier. 

\subsection{Case Studies (RQ4)}\label{sec:case-studies}
We conduct case studies to analyze the readability of the generated review summary and the sentiment consistency between the predicted sentiment labels and summaries. Table~\ref{tab:case-studies} compares the predicted sentiment labels and the generated summaries from the HSSC+copy model\footnote{HSSC+copy is a strong baseline which enhances the state-of-the-art model HSSC~\cite{DBLP:conf/ijcai/MaSLR18} with a copy mechanism~\cite{DBLP:conf/acl/SeeLM17}.} and our dual-view model on the testing sets. We use ``oracle'' to denote a model that always outputs the ground-truth. 
From row ``a'' to row ``d'' of Table~\ref{tab:case-studies}, we observe that the sentiment labels and the summaries predicted by the HSSC+copy model are not consistent with each other. For example, in row ``a'' of the table, the sentiment label predicted by HSSC+copy is 3, which indicates a neutral sentiment and matches the ground-truth. However, its generated summary (``not worth the money'') conveys a negative sentiment. 
In row ``d'' of the table, the HSSC+copy model generates a summary with a positive sentiment that is similar to the ground-truth, but it predicts a sentiment label of 3, which is not consistent with the positive sentiment of the generated summary. On the other hand, the sentiment labels and the review summaries predicted by our dual-view model are sentimentally consistent to each other on these rows. 
Moreover, the HSSC+copy model sometimes omits sentiment words in the generated summaries, which makes the summaries less informative. For example, in row ``f'' of the table, the summary generated by HSSC+copy omits the sentiment word ``paramount''. Thus, it does not reflect the opinion that the user likes the product very much, while the summary generated by our model contains ``paramount'' which expresses the consumer's opinion accurately. 


\begin{table}
\caption{Samples of the predicted sentiment labels and review summaries from different models. 
``Oracle'' denotes a model that always outputs the ground-truth. The predicted sentiment labels (boldfaced and italicized) are placed before each generated summary. The sentiment labels ranges from 1 to 5, where 5 (1) denotes the most positive (negative) sentiment attitude. 
}
\label{tab:case-studies}
\resizebox{\columnwidth}{!}{
\begin{tabular}{|c |c | p{0.68\columnwidth} |}
\hline 
\textbf{Id} & \textbf{Model} & \textbf{Predicted Sentiment Label and Summary} \\
\hline
\multirow{3}{*}{a} & Oracle & \textit{\textbf{3:}} you get what you pay for . \\
& HSSC+copy & \textit{\textbf{3:}} not worth the money . \\
& Dual-view & \textit{\textbf{3:}} you get what you pay for . \\
\hline
\multirow{3}{*}{b} & Oracle & \textit{\textbf{1:}} do not buy it . \\
& HSSC+copy & \textit{\textbf{1:}} this thing is okay . \\
& Dual-view & \textit{\textbf{1:}} not worth it . \\
\hline 
\multirow{3}{*}{c} & Oracle & \textit{\textbf{1:}} worst toaster i have ever owned .\\
& HSSC+copy & \textit{\textbf{2:}} worst toaster i have ever owned .\\
& Dual-view & \textit{\textbf{1:}} worst toaster i have ever owned .\\
\hline
\multirow{3}{*}{d} & Oracle & \textit{\textbf{5:}} best for side/stomach sleepers . \\
& HSSC+copy & \textit{\textbf{3:}} good for side/stomach sleepers . \\
& Dual-view & \textit{\textbf{5:}} comfortable for side/stomach sleepers . \\
\hline
\multirow{3}{*}{e} & Oracle & \textit{\textbf{5:}} makes great ice cream ! \\
& HSSC+copy & \textit{\textbf{5:}} ice cream maker . \\
& Dual-view & \textit{\textbf{5:}} great ice cream attachment ! \\
\hline
\multirow{3}{*}{f} & Oracle & \textit{\textbf{5:}} paramount stockton 5-piece daybed ensemble , twin leggett \& platt - home textiles . \\
& HSSC+copy & \textit{\textbf{5:}} 5-piece daybed ensemble . \\
& Dual-view & \textit{\textbf{5:}} paramount stockton 5-piece daybed ensemble . \\
\hline
\multirow{3}{*}{g} & Oracle & \textit{\textbf{4:}} works great for less money . \\
& HSSC+copy & \textit{\textbf{5:}} it 's a rangefinder .  \\
& Dual-view & \textit{\textbf{4:}} great for the price .\\
\hline
\end{tabular}
}
\end{table}

\section{Conclusion}
We propose a novel dual-view model with inconsistency loss to jointly improve the performance of review summarization and sentiment classification. 
Compared to previous work, our model has the following two advantages. First, it encourages the sentiment information in the decoder states to be similar to that in the review context representation, which assists the decoder to generate a summary that has the same sentiment tendency as the review. Second, the source-view and summary-view sentiment classifiers can learn from each other to improve the sentiment classification performance. 
Experiment results demonstrate that our model achieves better performance than the state-of-the-art models for the task of joint review summarization and sentiment classification. 


\begin{acks}
The work described in this paper was partially supported by the Research Grants Council of the Hong Kong Special Administrative Region, China (CUHK 2300174 (Collaborative Research Fund, No. C5026-18GF)). We would like to thank the four anonymous reviewers for their comments.
\end{acks}

\bibliographystyle{ACM-Reference-Format}
\bibliography{sample-base}



\end{document}